\documentclass[10pt,twocolumn,letterpaper]{article}

\usepackage{iccv}
\usepackage{times}
\usepackage{epsfig}
\usepackage{graphicx}
\usepackage{amsmath}
\usepackage{amssymb}

\usepackage[pagebackref=true,breaklinks=true,letterpaper=true,colorlinks,bookmarks=false]{hyperref}

\iccvfinalcopy % *** Uncomment this line for the final submission

\def\iccvPaperID{719} % *** Enter the ICCV Paper ID here

\ificcvfinal\fi
\begin{document}

%%%%%%%%% TITLE
\title{See the Glass Half Full: Reasoning about Liquid Containers,\\ their Volume and Content}

\author{Roozbeh Mottaghi$^\dagger$ \qquad Connor Schenck$^\ddagger$
\qquad Dieter Fox$^\ddagger$
\qquad Ali Farhadi$^{\dagger\ddagger}$\\
$^\dagger$Allen Institute for Artificial Intelligence (AI2)\\
$^\ddagger$University of Washington
}

\maketitle
\thispagestyle{empty}

%%%%%%%%% ABSTRACT
\begin{abstract}
Humans have rich understanding of liquid containers and their contents; for example, we can effortlessly pour water from a pitcher to a cup. Doing so requires estimating the volume of the cup, approximating the amount of water in the pitcher, and predicting the behavior of water when we tilt the pitcher. Very little attention in computer vision has been made to liquids and their containers. In this paper, we study liquid containers and their contents, and propose methods to estimate the volume of containers, approximate the amount of liquid in them, and perform comparative volume estimations all from a single RGB image. Furthermore, we show the results of the proposed model for predicting the behavior of liquids inside containers when one tilts the containers. We also introduce a new dataset of Containers Of liQuid contEnt (COQE) that contains more than 5,000 images of 10,000 liquid containers in context labelled with volume, amount of content, bounding box annotation, and corresponding similar 3D CAD models. 
\end{abstract}

%%%%%%%%% BODY TEXT
\section{Introduction}

Recent advancements in visual recognition have enabled researchers to start exploring tasks that go beyond categorization and entail high-level reasoning in visual domains. Visual reasoning, an essential component for a visually intelligent agent, has recently attracted computer vision researchers~\cite{wu15,newtonian,mottaghi16,lerer16,pham15,zhu15,zhu16,brubaker09,agrawal16,li16}. Almost all the efforts in visual recognition and reasoning have been devoted to solid objects: how to detect~\cite{fastrcnn,fasterrcnn} and segment ~\cite{long15,chen15} them, how to reason about physics of a solid world \cite{newtonian, wu15}, and how forces would affect their behavior~\cite{mottaghi16,lerer16}. Very little attention, however, has been made to liquid containers and the behavior of their content. 

Humans, on the other hand, deal with liquids and their containers on a daily basis. We can comfortably pour water from a pitcher to a cup knowing how much water is already in the cup and having an estimate of the volume of the cup and the amount of water in the pitcher. We effortlessly estimate the angle by which to tilt the pitcher to pour the right amount of water into the cup. In fact, five month old infants develop rich understanding of liquids and their containers and can predict whether water will pour or tumble from a cup if the cup is upended~\cite{hespos09}. Other species such as orangutans can also estimate the volume of liquids inside a container and can predict if the liquid in one container can fit into the other one~\cite{call97}.

\begin{figure}[t]
\begin{center}
\includegraphics[width=20.5pc]{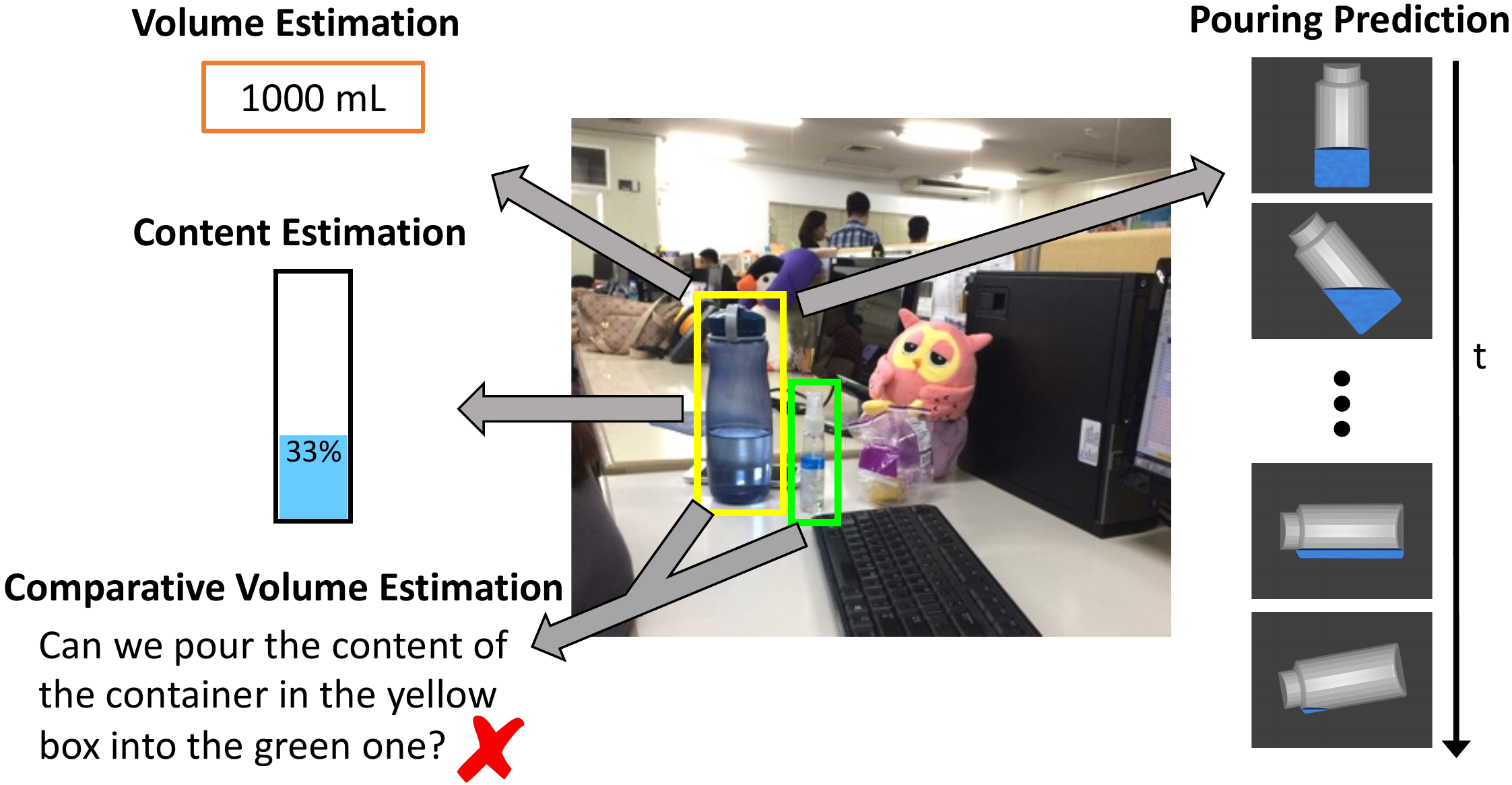}
\caption{Our goal is to estimate the volume of the container (Volume Estimation), approximate what fraction of the volume is filled (Content Estimation), infer whether we can pour the content of one container into another (Comparative Volume Estimation), and predict how much liquid will remain in a container over time if it is tilted to a certain angle (Pouring Prediction). Our inference is based on a single RGB image.}
\label{fig:intro}
\end{center}
\end{figure}

In this paper, we study liquid containers (Figure~\ref{fig:intro}) and propose methods to estimate the volume of containers and their content in absolute and relative senses. We also show, for the first time, that we can predict the amount of liquid that remains in a container if it is tilted for a certain tilt angle, all from a single image. We introduce Containers Of liQuid contEnt (COQE), a dataset of images with containers annotated for their volume, the volume of their content, bounding box, and corresponding similar 3D models. Estimating the volume of containers is extremely challenging and requires reasoning about the size of the container, its shape, and contextual cues surrounding the container. The volume of the liquid content of a container can be estimated using subtle visual cues like the line at the edge of the liquid inside the container. We propose a deep learning based method to estimate the volume of containers and their content using the contextual cues from the surrounding objects. In addition, by integrating Convolutional Neural Networks and Recurrent Neural Networks, we can predict the behaviour of liquid contents inside containers as their reaction to tilting the container. 

Our experimental evaluations on COQE dataset show that incorporating contextual cues provides improvement for estimating volume of the containers and the amount of their content. Furthermore, we show the results using a single RGB image for predicting how much liquid will remain inside a container over time if it is tilted by a certain angle. %Generally, it is interesting that off-the-shelf state-of-the-art CNNs do not perform reliably on these tasks, which illustrates how challenging these tasks are. 

\section{Related Work}
In this section, we describe the work relevant to ours. To the best of our knowledge, there is little to no work that directly addresses the same problem. Below, we mention past work that are most related. 

In \cite{fritz09}, a hybrid discriminative-generative approach is proposed to detect transparent objects such as bottles and glasses. \cite{phillips16} propose a method for detection, 3D pose estimation, and 3D reconstruction of glassware. \cite{ye15} also propose a method for reconstruction of 3D scenes that include transparent objects. Our work goes beyond detection and reconstruction since we perform reasoning about higher-level tasks such as content estimation or pour prediction. 

Object sizes are inferred by \cite{hessam16} using a combination of visual and linguistic cues. In this paper, we focus only on visual cues. Size estimates have also been used by \cite{hoiem06, stark12} to better estimate the geometry of scenes. %In this paper, we address the volume rather than size.

The result of 3D object detectors \cite {lin13,song14,gupta15} can be used to obtain a rough estimate of the volume of the containers. However, they are typically designed for RGBD images. Moreover, the output of these detectors cannot be used for estimation of the amount of content or pouring prediction. Depth estimation methods from single RGB images \cite{delage06,liu10,eigen14} can also be used for computing the relative size of containers. 

\begin{figure*}[tp!]
\begin{center}
\includegraphics[width=36pc]{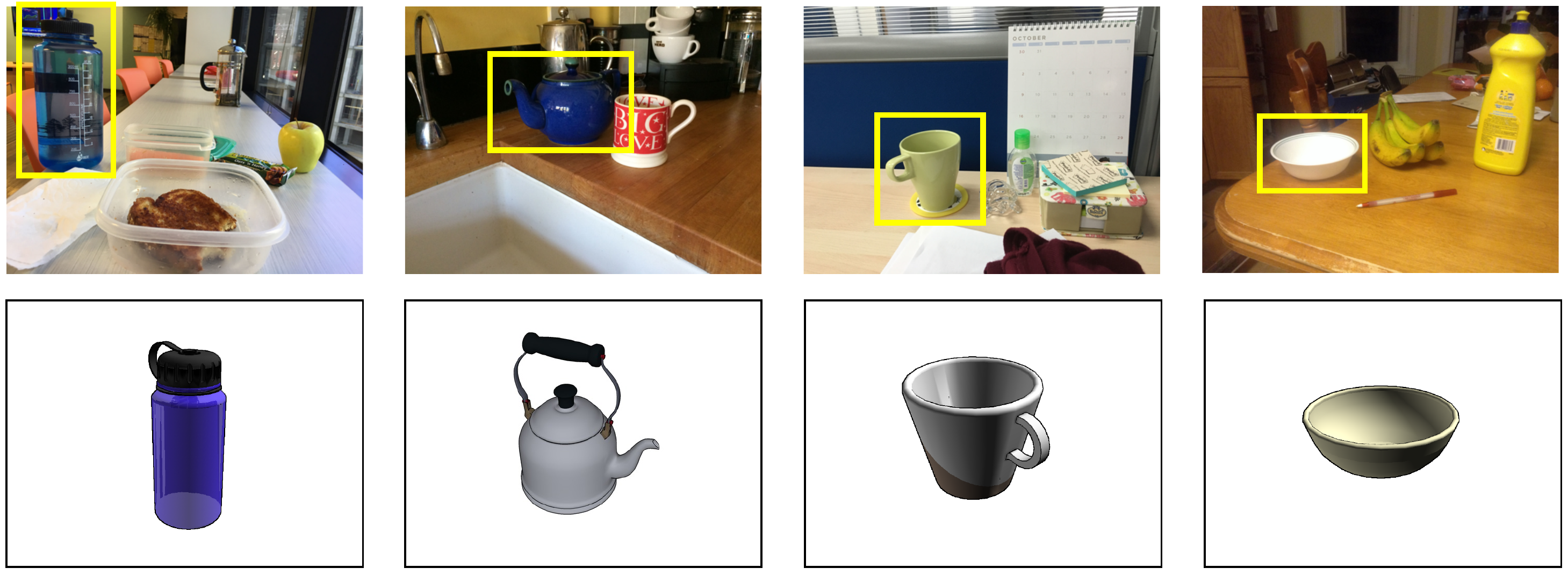}
\caption{\textbf{COQE dataset.} Example containers in our dataset. The bottom row shows the corresponding 3D CAD model for the container inside the yellow bounding box.}
\label{fig:dataset}
\end{center}
\end{figure*}

The affordance of containing liquids is inferred by \cite{yu15}. Additionally, they reason about the best filling and transfer directions. The problem that we address is different and we use RGB images during inference (as opposed to RGBD images). \cite{liang15} uses physical simulation to infer the affordance of containers and containment relationship between objects. Our work is different since we reason about liquid content estimation, pouring prediction, etc. 

Our pouring prediction task shares similarities with \cite{mottaghi16}. In \cite{mottaghi16}, they predict the sequence of movement of rigid objects for a given force. In this work, we are concerned with liquids that have different dynamics and appearance statistics than solid objects. 

There are a number of works in the robotics community that tackle the problem of liquid pouring \cite{tamosiunaite11, Brandl14, rozo13,yamaguchi15,kunze15,schenck16}. However, these approaches either have been designed for synthetic environments \cite{yamaguchi15,kunze15} or they have been tested in lab settings and with additional sensors \cite{tamosiunaite11, rozo13,Brandl14,schenck16}. Fluid simulation is a popular topic in computer graphics \cite{muller03,kim08,bridson15}. Our problem is different since we predict the liquid behavior from a single image and are not concerned about rendering. 

There are also several cognitive studies about liquids, their physical properties and their interaction with the containers \cite{collins87,schwartz99,cohn01,hespos07,Bates15,strickland15,strickland15,kubricht16,liang16}. 

\section{Tasks}
In this paper, we focus on four important tasks related to liquids and their containers:
\begin{itemize}
\item \textbf{Container volume estimation:} Our goal in this task is to infer the volume of the container (i.e, the volume of the liquid inside the container when the container is full). The input is a single RGB image and the query container, and the output is the volume estimate (e.g., 50mL, 200mL, etc).
\item\textbf{Content estimation:} In this task, the goal is to estimate how full a container is given a single RGB image and a query container. The example outputs are \emph{empty}, 10\% \emph{full}, 50\% \emph{full}, etc.

\item \textbf{Comparative volume estimation:} The task is to infer if we can pour the entire content of one container into another container. The input is a single RGB image and a pair of query containers in that image, and the output is \emph{yes}, \emph{no}, or \emph{can't tell} (since we have opaque containers in the dataset). This is more complex than the previous two tasks since it requires reasoning about the size of the two containers and the amount of liquid in them simultaneously. 
\item \textbf{Pouring prediction:} The goal is to infer the amount of liquid in a container over time after tilting the container by a given angle. The inputs are a single RGB image, a query object, and a tilt angle. The output is a variable length sequence that determines the amount of liquid at each time step. The sequence has a variable length since some containers become empty much faster than other containers depending on the initial amount of liquid in them, the size of the container, and the tilt angle.
\end{itemize}

\section{COQE Dataset}
\label{sec:dataset}
There is no dataset to train and evaluate models on the four tasks defined above. Hence, we introduce a new dataset called Containers Of liQuid contEnt (COQE).

The COQE dataset includes more than 5,000 images, where in each image there are at least two containers. The containers belong to different categories such as bottle, glass, pitcher, bowl, kettle, pot, etc. Figure~\ref{fig:dataset} shows some example images in the dataset.

It is infeasible to use web for collecting this dataset since obtaining accurate groundtruth volume estimates for arbitrary web images is not trivial. To overcome this problem, we used a commercial crowd-sourcing platform to collect images and their corresponding annotations. The annotators took pictures using their cameras or cellphones and measured the container volume using a measuring cup or reported the volume on the container label. 

%The collected images follow a set of requirements so we can better evaluate the mentioned tasks.
The data collectors were instructed to meet certain requirements. First, the images should include the context around the container since estimating the size from an image that only shows the container is an ambiguous task. To impose this constraint, we asked the annotators to take pictures that have at least 4 objects in each image. Second, the dataset should include annotations only for containers that had a bounding box whose larger side is larger than 30 pixels. We had this requirement because the content of the containers is not visible if the containers appear very small in the image. Finally, the dataset should include images that have objects in a natural setting to better capture background clutter, different illumination conditions, occlusion, etc. 

Each container in our dataset has been annotated by its bounding box, the volume, and the amount of liquid inside the container. Additionally, we downloaded 34 CAD models from Trimble 3D Warehouse and we specify which 3D CAD model is most similar to each container in the images. Finding the correspondence with the CAD models enables us to run pouring simulations. For pouring simulation, we rescale the CAD models to the annotated volume and consider the annotated amount of liquid in the CAD model. Then, we tilt the CAD model by $x$ degrees and record how much liquid remains in the CAD model for each tilt angle. Section~\ref{sec:exppour} provides more details about pouring simulations.

% \begin{figure*}[tp!]
% \begin{center}
% \includegraphics[height=8pc]{net_arch.pdf}
% \caption{}
% \label{fig:pouring}
% \end{center}
% \end{figure*}

\section{Our Approach}
%In this section, we explain the models for the proposed tasks. First, we describe our model for volume and content estimation. Then, we will show how we perform pairwise comparisons. Finally, we describe the sequence generation model for pouring liquids.

\subsection{Volume and Content Estimation}
\label{sec:modelvolume}
We now describe the model for estimation of   \emph{container} volume and \emph{content} volume for a query container in an image. 

We use a Convolutional Neural Network (CNN), where the input has 4 channels. The first three channels of the input are the RGB channels of the input image, and the fourth channel is used to represent the bounding box of the query container, which is basically a bounding box mask smoothed by a Gaussian kernel. An additional input to our model is a set of masks generated by an object detector. The masks generated by the object detector enable us to capture contextual information. The idea is that the surrounding objects typically provide a rough prior for the volume of the container of interest. We use Multipath network \cite{zagoruyko16} as our object detector, which is a state-of-the-art instance segmentation method that generates a mask for objects along with the category information. We use Multipath that is trained on COCO dataset \cite{lin14} so it generates masks for 80 categories defined by \cite{lin14}. We create a binary image for each category, where the pixels of all masks for that category are set to 1. Then, we resize the mask to $28\times28$. We obtain a $28\times28\times81$ cube, referred to as \textit{context tensor}, since the object detector has 80 categories and we consider one category for the background (areas not covered by the masks of the 80 categories). For efficiency concerns, we do not use these masks in the input channel and we use them in a higher level of the model. 

\begin{figure}[tp!]
\begin{center}
\includegraphics[width=20pc]{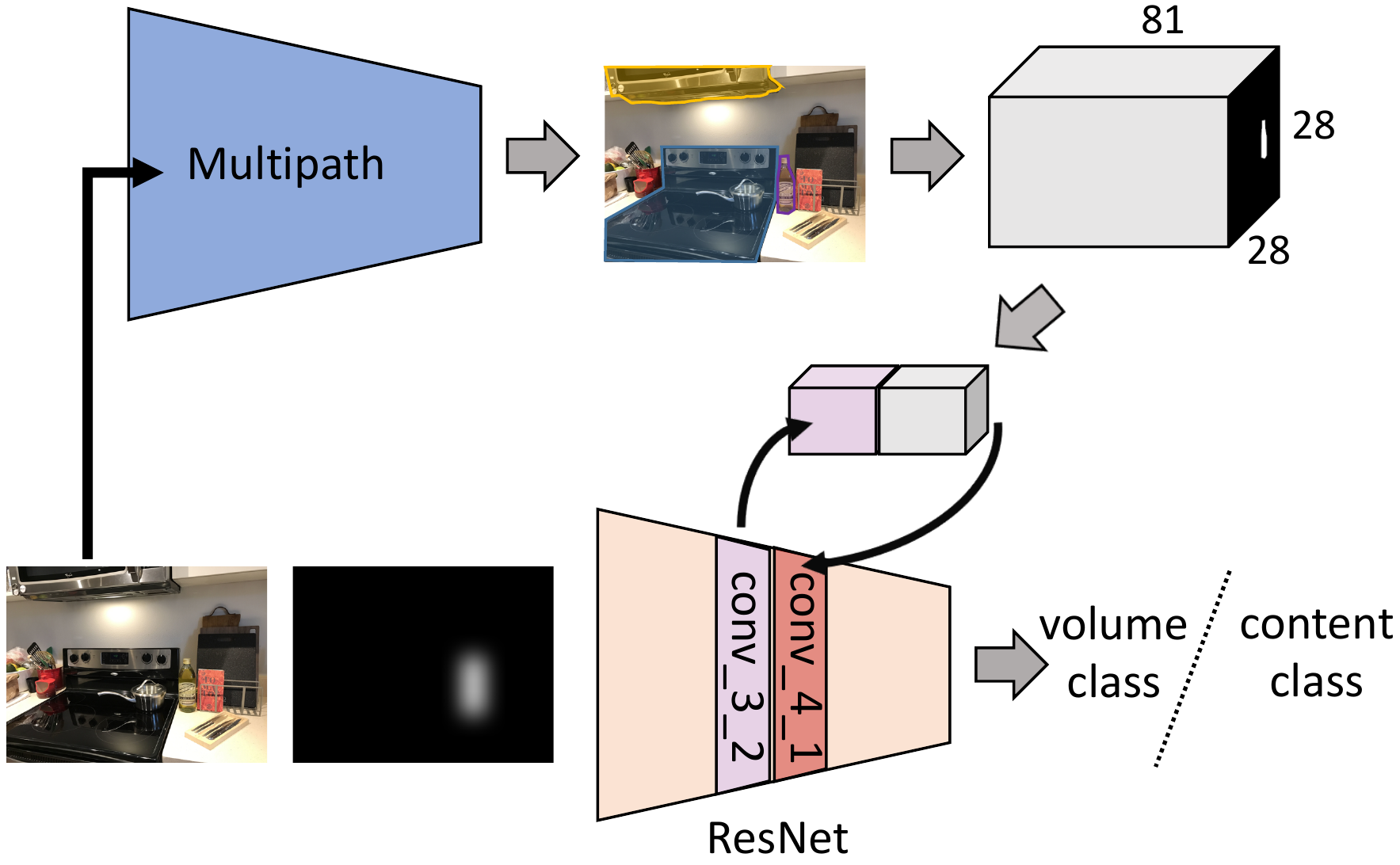}
\caption{\textbf{Model architecture.} For volume and content estimation, the input to our network is an RGB image and the mask for the query container. We feed the RGB image into the Multipath network \cite{zagoruyko16} to generate a set of mask detections. The masks for different categories form a tensor (shown in grey) and are concatenated with the output of the \texttt{conv\_3\_2} layer of ResNet-18 (the purple cube).}
\label{fig:model}
\end{center}
\end{figure}

The architecture of our model is shown in Figure~\ref{fig:model}. We concatenate the context tensor with the input of the \texttt{conv4\_1} layer of ResNet-18 \cite{resnet} whose input size is $28\times 28\times 128$. As a result, the input to \texttt{conv4\_1} will be of size $28\times 28\times 209$. We refer to this network as Contextual ResNet for Containers (CRC) throughout the paper. 

We formulate volume and content estimation as classification. We change the layer before the  classification layer of ResNet based on the number of classes in each task. The loss for this network is the cross-entropy loss, which is typically used for multi-class classification. We consider different weights for different classes according to their inverse frequency in the training data. We could alternatively formulate these tasks as a regression problem. However, we obtained better performance using the classification formulation. Note that we train the network separately for volume and content estimation tasks (i.e. the classification layer has different size of output depending on the task). 

%We convert the detections to a $28\times28\times81$ cube, where the third dimension corresponds to 80 categories plus background. 

\subsection{Comparative Volume Estimation}
 Here, we answer the following question: ``Can we pour the entire content of container 1 into container 2 in the same image?". Basically, the model needs to estimate the volume for the two containers and infer the current amount of liquid in each of them to answer the question. Our approach is implicit in that we let the network figure out these quantities and do not provide explicit supervision. 

Our model for this task is a Siamese network, where there are two branches corresponding to two different containers in question. Similar to the previous tasks, each branch of the model receives a 4-channel input, where the first 3 channels is the RGB image and the 4th channel is the bounding box mask for the query container. We concatenate the output of the layers before the classification layers of the two branches (the concatenation output is 1024-dimensional). A fully connected (FC) layer follows the output of the concatenation, which provides the input to a Log-Softmax layer. Alternatively, we tried a 5-channel input (i.e., 3 RGB channels, one channel for the mask of container 1 and another channel for the mask of container 2). The performance for this scenario was worse than the performance of the proposed model. We also tried two scenarios for the Siamese network, where we considered shared and non-shared weights. The performance for the shared weight case was better. The loss for this task is cross-entropy loss as well since we formulate it as classification, where the labels are \emph{yes}, \emph{no}, \emph{can't tell} (which happens when at least one of the containers is opaque and its content is not visible). 

% \begin{figure}[h]
% \begin{center}
% \includegraphics[width=0.5\linewidth]{net_arch.pdf}
% \caption{}
% \label{fig:model_pair}
% \end{center}
% \end{figure}

\subsection{Pouring Prediction}
\label{sec:pourpredmodel}
In this task, we predict how much liquid will remain in the container if we tilt it by $x$ degrees. The output of this task is a function of a few factors: (1) The initial amount of liquid in the container, e.g., if a bottle is 10\% full, tilting it by a few degrees will not have any effect on the amount of the liquid that remains inside the container. (2) The geometry of the container. For example, a large tilt angle is required to pour the liquid from a container that has a narrow mouth. (3) The volume of the container. For example, it takes longer to pour the content of a larger container compared to a tiny container. Estimating each of these factors is a challenging task by itself. 

We formulate this task as sequence prediction, where our goal is to generate the sequence of the amount of liquid in the container over time given a single RGB image, a query container, and a tilt angle $x$. 

The amount of the liquid at each time step is dependent on the previous time steps so we use a recurrent network to capture these dependencies. Our architecture is a Convolutional Neural Network (CNN) followed by a Recurrent Neural Network (RNN).% similar to the architecture that is generally used for image captioning \cite{vinyals15,karpathy15}. 

\begin{figure}[tp]
\begin{center}
\includegraphics[width=20pc]{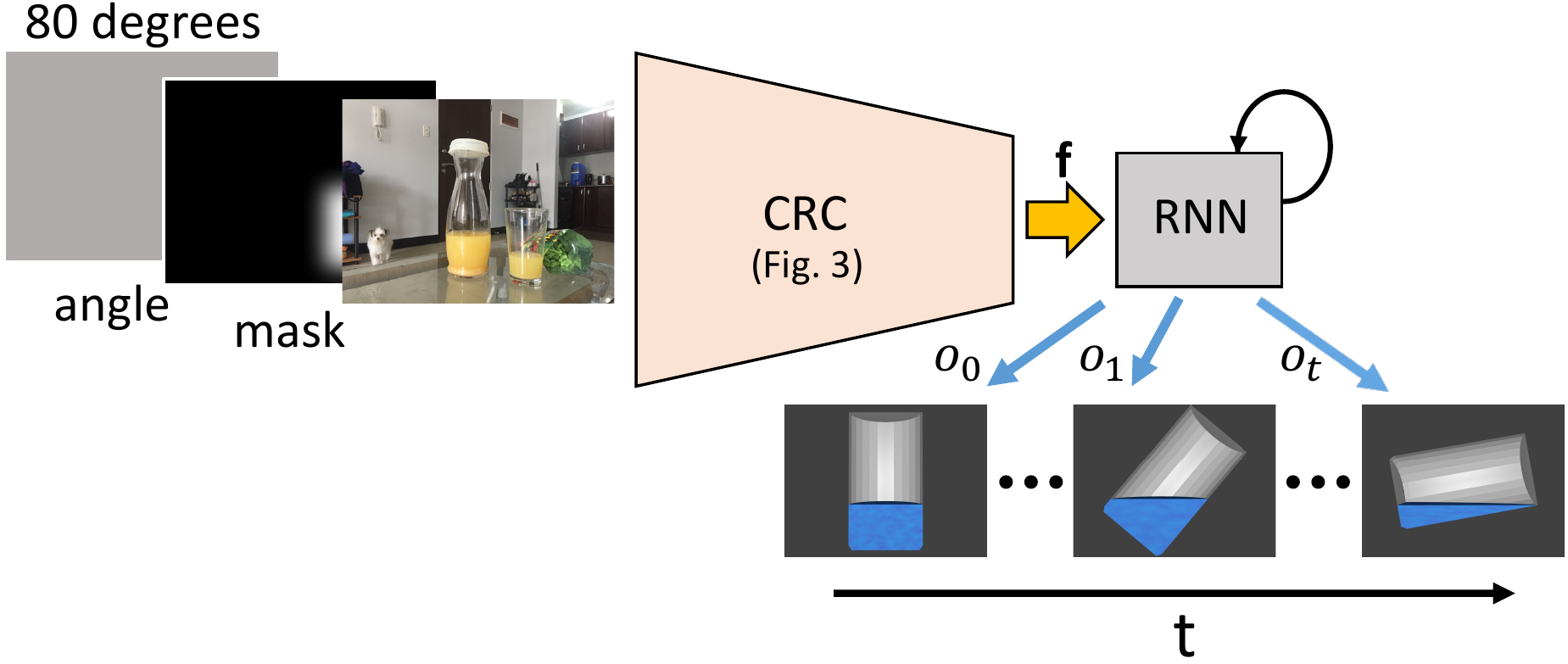}
\caption{\textbf{Model architecture for pouring prediction.} The input to this model is an RGB image, the mask for the query container and an image that encodes the tilt angle. The output of our model (CRC) is fed into an RNN that predicts a sequence that represents how much liquid remains in the container over time. We train this network end-to-end.}
\label{fig:model_rnn}
\end{center}
\end{figure}

The CNN part of the network has the same architecture as that of CRC (shown in Figure~\ref{fig:model}) with two differences. The first difference is that we have an additional input channel to encode the angle $x$. This channel has the same height and width as the input image and it is concatenated with the input image. All elements of this channel are set to $x$. The second difference is that we remove the classification layer of CRC so we can feed the output of the CNN into the recurrent part. We denote the output of the CNN by $\mathbf{f}$, which is a 512 dimensional vector. We use $\mathbf{f}$ as the input to the recurrent part of the network. The architecture for this task is shown in Figure~\ref{fig:model_rnn}. We consider a 100-dimensional hidden unit for the recurrent network. The output of the RNN at each time step, $\mathbf{o}_t$, is $|\mathcal{R}|$ dimensional, where $\mathcal{R}=\{r_0, r_1, \cdots, r_N, p\}$ is the set of discretized amounts of liquid, for example, $r_0$ represents \emph{empty}, $r_1$ represents 10\% full, etc. The label $p$ represents the opaque case. Some of the containers are opaque. In this case, no estimation can be provided since the content is not visible. Note that the problem at each time step is a classification problem, where the RNN generates one of $|\mathcal{R}|$ classes. As described in Section~\ref{sec:dataset}, there is a 3D CAD model associated to each example. Therefore, we can simulate tilting for each container given an initial amount of liquid and obtain the groundtruth for this task. Note that the 3D CAD models are only used during training and not for inference.The detailed procedure for obtaining the groundtruth sequence is described in Section~\ref{sec:exppour}.

The RNN stops the sequence if it generates $r_0$, which is the \emph{empty} state, or $p$, which corresponds to the \emph{opaque} container case. The reason is that the rest of the sequence should be the same if it generates either of these two labels. We consider a maximum length of 5 for the sequences in our experiments. 

The loss function is defined over the output sequence. Suppose we denote the groundtruth and output sequence by $S=(s_0, s_1,\cdots, s_{t'})$ and $O=(\mathbf{o}_0, \mathbf{o}_1,\cdots, \mathbf{o}_t)$, respectively. The loss will be defined as:
\begin{equation}
     L(S, O) = -\frac{1}{T}\sum_{t = 0}^T w_t(s_t)\,\, \mbox{log}(\mathbf{o}_t[s_t]),
\end{equation}
where $T$ is the maximum length of sequence, and $w_t(s_t)$ is the weight for each class (i.e. the inverse frequency of the amount $s_t$ at time $t$ in the training data). Also, $\mathbf{o}_t[s_t]$ is the $s_t$-th element of $\mathbf{o}_t$. Recall that $\mathbf{o}_t$ is $|\mathcal{R}|$-dimensional. Also, note that $\mathbf{o}_t = SoftMax(g(h_t))$, where $h_t$ is the hidden unit of the RNN at time step $t$, and $g$ is a linear function followed by a ReLU non-linearity. Hence, the loss is a cross-entropy loss defined over the sequence. If the output sequence and the groundtruth sequence have different lengths (i.e. $t\neq t'$) , we pad them by the last element of the sequence to make them the same length.

\section{Experiments}
We evaluate our models on different tasks that we defined: estimating the volume of a query container, estimating how full the container is (content estimation), comparative volume estimation that infers if we can pour the entire content of one query container into another, and pouring prediction that provides a temporal estimate of how much liquid will remain inside the query container if we tilt it. The first three tasks are mainly related to estimating the geometry of the container and its content, while the fourth task addresses the estimation of the behavior of the liquid inside the container. 

\noindent \textbf{Dataset:} Our dataset consists of more than 5,000 images that include more than 10,000 annotated containers. We use 6,386 containers for training, 1,000 for validation and 3,000 for test. Each container is annotated with the volume, the amount of content, a bounding box, and a corresponding 3D CAD model. %These annotations are used for training the models for different tasks. 

\subsection{Implementation Details}

We use Torch\footnote{http://torch.ch} to implement the proposed neural networks. We run the experiments on a Tesla K40 GPU. We feed the training images into the network in batches of size 96, where each batch contains RGB images, the mask images for the query container (or two masks for the comparative volume estimation task), and context tensor (described in Section~\ref{sec:modelvolume}). %We resize the input images and the mask to $224\times224$. For training our model and the baselines, we consider 20 random crops of each input image to augment our training data. However, we make sure that each random crop contains at least 50\% of the bounding box mask for the query container. 

Our learning rate is $10^{-3}$ for all experiments. We use ResNet-18 \cite{resnet} for the ResNet part of the networks. The ResNet is pre-trained on ImageNet \footnote{https://github.com/facebook/fb.resnet.torch/tree/master/pretrained}. We randomly initialize the mask channels of the input layer and additional channels of \texttt{conv\_4\_1} in CRC. For the random initialization, we randomly sample from a Gaussian distribution with mean 0 and standard deviation 0.01. To train the proposed models and the baselines we use 20,000 iterations. We choose the model that has the highest performance on the validation set.

\begin{figure*}[tp!]
\begin{center}
\includegraphics[width=40pc]{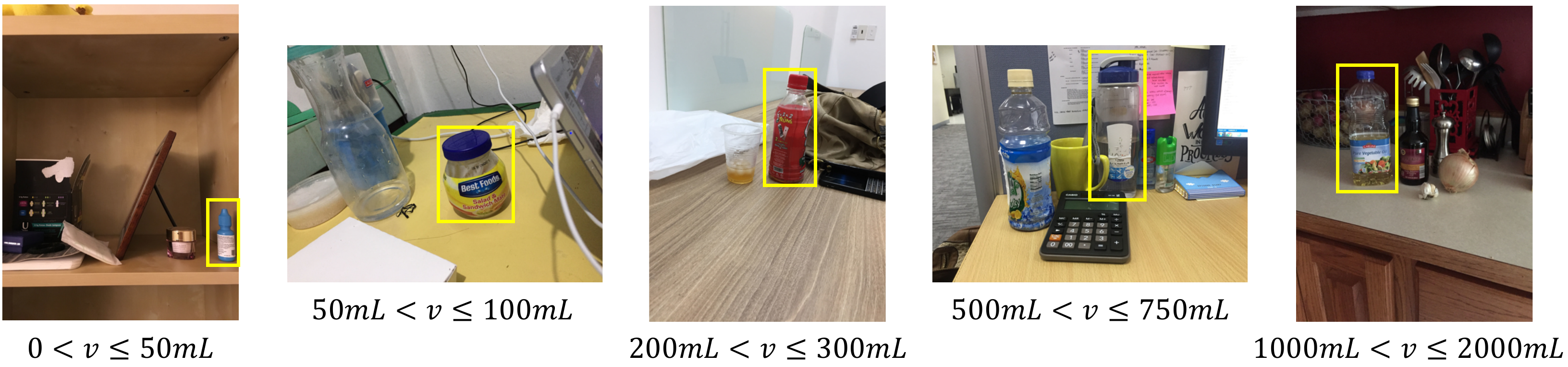}
\caption{\textbf{Qualitative results of volume estimation.} The volume for the query container (indicated by the yellow bounding box) is shown under the image.}
\label{fig:resvol}
\end{center}
\end{figure*}

\subsection{Volume Estimation}
We first provide evaluations for the volume estimation task. We divide the space of volumes into 10 classes, where the maximum volumes in each class are: 50, 100, 200, 300, 500, 750, 1000, 2000, 3000, $\infty$. The unit for the measurement is milliliter (mL). For example, the first class contains all containers that are smaller than 50mL, the second class are containers whose volume is between 50mL and 100mL and so on. The reason that the range is not uniform is to have better visual separation of examples. We could alternatively formulate the problem as a regression problem since volume is a continuous quantity, but the performance was worse. \cite{wang15,walker15, mottaghi16} also formulated a continuous variable estimation problem as classification due to the same reason. 

The baselines for this task are: (1) a naive regression that takes width and height of the container bounding box (normalized by the image width and height, respectively) as features and regresses the volume. (2) classification using AlexNet, where we replace the FC7 layer of AlexNet and its classification layer to adapt them to a 10-class classification. (3) The CRC model without the contextual information. We use the same number of iterations for training these networks. 

Table~\ref{tab:volume} shows the results for this task. Our evaluation metric is average per-class accuracy. The chance performance for this task is 10\%. Our model provides about 2.5\% improvement over the case that we do not use contextual information. The results suggest that the information about the surrounding objects can help volume estimation. The overall low performance of these state-of-the-art CNNs shows how challenging the task is. Figure~\ref{fig:resvol} shows qualitative examples of volume estimation.

\begin{table}[h]
\setlength{\tabcolsep}{3pt}
\begin{center}
\begin{tabular}{c|c}
    \hline
    & Avg. per-class accuracy \\
    \hline
        Chance & 10.00 \\
        Box-Regression & 11.36 \\ 
        AlexNet &  17.33 \\
        Ours w/o context & 15.33  \\
        Ours w/ context (CRC) & \textbf{17.79}\\
    \hline
\end{tabular}
\end{center}
\caption{Quantitative results for \textbf{volume estimation}.}
\label{tab:volume}
\end{table}

\subsection{Content Estimation}
In this task, we estimate the amount of content in a query container in an RGB image. The annotators provide groundtruth for this task in terms of one of the following 6 classes: 0\% (empty), 33\%, 50\%, 66\%, 100\%, and opaque. The content of an opaque container cannot be estimated using visual cues so we consider this category as well to handle this case. Similar to above, we use average per-class accuracy as the evaluation metric. We use similar CNN-based baselines as above.

%As the baseline, we modify AlexNet and ResNet for performing classification on this task (similar changes as above, but for a different number of classes). 

\begin{figure*}[tp!]
\begin{center}
\vspace{-0.1cm}
\includegraphics[width=39pc]{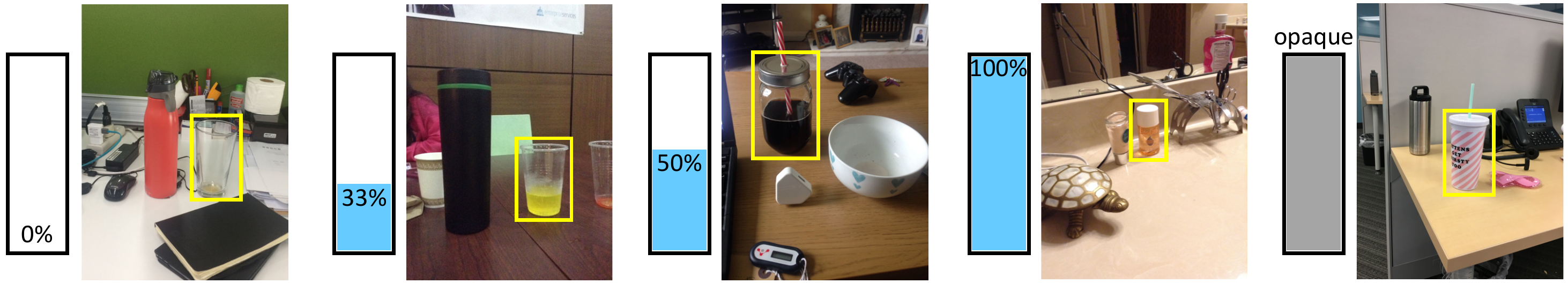}
\caption{\textbf{Qualitative results of content estimation.} On the left side of each image, we show the predicted amount of liquid in the query container (indicated by the yellow box). The rightmost image shows an opaque container for which it is not possible to correctly predict the amount of content.}
\vspace{-0.6cm}
\label{fig:rescontent}
\end{center}
\end{figure*}

Table~\ref{tab:content} shows the results for this task. Our model improved the performance by 2.7\% compared to the case that we do not use context. The chance accuracy for this task is 16.67\%. Our method achieves 32.01\% per-class accuracy which is 15.3\% above the chance performance. However, this result shows that there is still a large room for improvement on this task. Figure~\ref{fig:rescontent} shows a few qualitative examples of content estimation.

\begin{table}[h]
\setlength{\tabcolsep}{3pt}
\begin{center}
\begin{tabular}{c|c}
    \hline
    & Avg. per-class accuracy \\
    \hline
        Chance & 16.67 \\
        AlexNet &  29.30 \\
        Ours w/o context & 29.29  \\
        Ours w/ context (CRC) & \textbf{32.01}\\
    \hline
\end{tabular}
\end{center}
\caption{Quantitative results for \textbf{content estimation}.}
\vspace{-0.3cm}
\label{tab:content}
\end{table}

\subsection{Comparative Volume Estimation}
In this task, we infer whether we can pour the \emph{entire} content of a query container into another one. This is a challenging task since it requires estimation of the content volume for both containers and also the volume of the container that the liquid is poured into. We formulate this problem as a 3-class classification, where the classes are \emph{yes}, \emph{no}, and \emph{can't tell} (when at least one of the containers is opaque). 

The procedure for obtaining groundtruth for this task is as follows. Let $v_1$ and $v_2$ to be the volume for a pair of containers in an image, respectively, and $c_1$ and $c_2$ represent how full each container is ($0\leq c_1,c_2\leq1$). Note that in our dataset we have annotations for $v_1$, $v_2$, $c_1$, and $c_2$. If $c_1 * v_1 < (1-c_2) * v_2$, we can pour the entire content of container 1 into container 2. 

For this experiment, two 4-channel input images are fed into the two branches of the Siamese network. As baselines, we replace both branches of the network by AlexNets or our model without context, where a fully connected (FC) layer and a Log-Softmax layer follow the concatenation of the output of these branches. Similar to the previous tasks, we use average per-class accuracy as the evaluation metric.

Table~\ref{tab:pairwise} shows the results. Note that in this task, we consider only containers that are in the same image since comparative volume estimation across different images is a difficult task even for humans. %The qualitative results for this task is shown in Figure~\ref{fig:respour}.

% \begin{figure*}[tp]
% \begin{center}
% \includegraphics[width=40pc]{figs/pour-old}
% \caption{}
% \label{fig:respour}
% \end{center}
% \end{figure*}

\begin{table}[h]
\setlength{\tabcolsep}{3pt}
\begin{center}
\begin{tabular}{c|c}
    \hline
    & Avg. per-class accuracy \\
    \hline
        Chance & 33.33 \\
        AlexNet &  43.90 \\
        Ours w/o context & 48.97  \\
        Ours w/ context (CRC) & \textbf{49.81}\\
    \hline
\end{tabular}
\end{center}
\caption{Quantitative results for \textbf{comparative volume estimation}.}
\vspace{-0.4cm}
\label{tab:pairwise}
\end{table}

\subsection{Pouring Prediction}
\label{sec:exppour}
The above evaluations mainly address the properties of the containers such as the volume of the containers and the amount of their content. We now describe the results for pouring prediction task, which is related to the behavior of the liquid inside the containers. This task requires generating a sequence, where each element of the sequence shows how full the container is at each timestep. We first explain how we obtain groundtruth sequences and then present the evaluations.  

\begin{figure*}[t]
\begin{center}
\includegraphics[width=38pc]{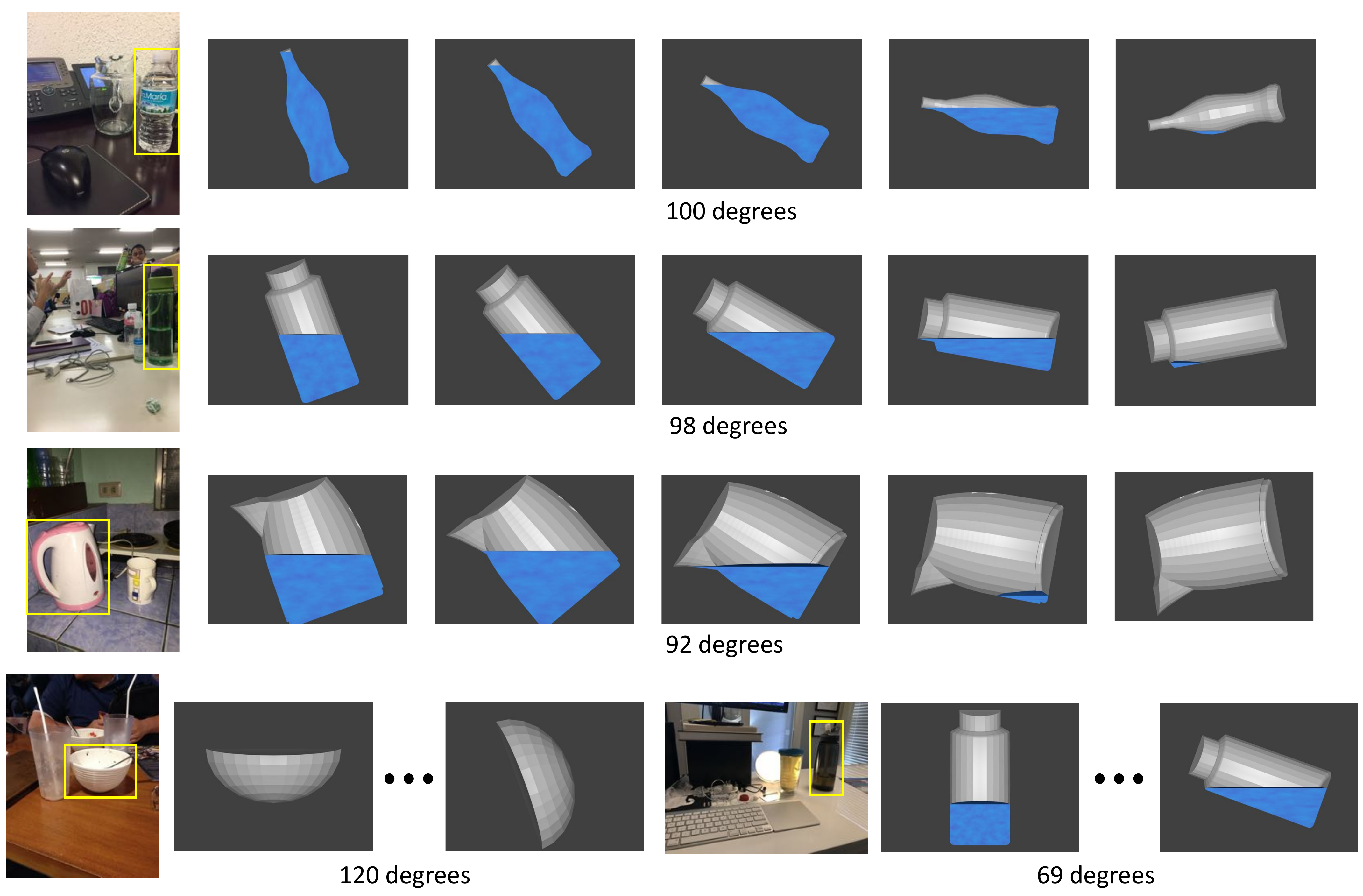}
\caption{\textbf{Qualitative results for pouring prediction.} Our method estimates the amount of the remaining liquid at each time step. The tilt angle for each sequence is shown under the sequence. The bottom row shows the case that the amount of liquid in the container does not change as the result of tilting. Note that we show the CAD models only for visualization purposes. They are neither predicted nor used for inference.}
\label{fig:resangle}
\end{center}
\end{figure*}

\noindent \textbf{Obtaining groundtruth:} Recall that we have a 3D CAD model associated to each container in images. Therefore, we can consider a certain amount of liquid in each 3D CAD model. We compute the amount of liquid remaining at each timestep during a pour as follows. At each timestep, we use the angle of the container to compute the maximum amount of liquid that could stably be held in the container without overflowing. To do this, we draw a horizontal plane parallel to the ground from the lip of the container. We then compute the volume of the container below that plane using a 3D mesh of its interior. Then, to compute the remaining amount of liquid at that timestep, we simply take the maximum of this value and the initial amount of liquid in the container. Intuitively, this means that if the container at a given angle can hold more liquid than it was initially filled with, then none will have spilled out and that is the amount that is in the container. Conversely, if the maximum amount of liquid that can rest stably in a container is less than the initial amount, then all excess will have spilled out and the amount remaining will be the maximum stable amount\footnote{Note that this approximation does not take into account attributes such as liquid viscosity or surface tension. However, this approximation is accurate enough for our purposes.}. We also used Fluidix (a fluid simulation library), but the results were not significantly different from the results of the above method (the error was smaller than our bin size).

During training, for each container in the images, we have an associated CAD model and an initial amount of liquid in the container (one of the following values according to the annotations: 0\%, 33\%, 50\%, 66\%, 100\%, or opaque). Therefore, we can estimate the amount of remaining liquid in the container for different angles and different initial amount of liquid. Note that during test we only have a single RGB image, the mask for the query container and the query angle, and we do not use 3D CAD models.

To generate sequences, we tilt each container from 0 degrees to a certain degree $x$, where 0 degree is the upright pose and 180 degrees corresponds to an upside down container (we ignore containers that are not in the upright pose in the image for training and evaluation). The maximum length of sequence that we consider is 5 i.e. we consider 5 timesteps for tilting from 0 to $x$ degrees and measure the remaining amount of liquid at each timestep using the procedure described above. We consider a discrete set of fractions $0, 0.1, 0.2,\ldots, 0.9, 1$ and assign the remaining amount of liquid to the closest fraction. Therefore, each element of the sequence belongs to one of 12 classes (11 fractions + 1 opaque class). More concretely, in $\mathcal{R}=\{r_0, r_1, \cdots, r_N, p\}$ (defined in Section~\ref{sec:pourpredmodel}), $r_0=0$, $r_1=0.1$, $r_2=0.2$, etc.

Note that the sequences can have different length. For example, if a container is initially empty, the sequence will be of length 1 since the amount of liquid will not change as the result of tilting. Similarly, the corresponding sequence for all opaque containers is of length 1 since no estimation can be performed for opaque containers. 

The result for this task is shown in Table~\ref{tab:pourpred}. Our evaluation criteria is defined as follows. We consider a predicted sequence as correct if all elements of that sequence match the elements in the groundtruth sequence. The first column of the table shows the result of the exact match of the sequences. We also show the results for different edit distances (edit distance between the predicted and groundtruth sequences). Qualitative examples of pouring prediction are shown in Figure~\ref{fig:resangle}.

We apply 5 different tilt angles to each container in train, validation and test images. The chance performance for this task is $1/192$ since there are 192 unique patterns of sequences in the test set. 

\begin{table}[h]
\setlength{\tabcolsep}{3pt}
\begin{center}
\begin{tabular}{|c|c|c|c|c|c|c|}
    \hline
    Edit distance & 0 & 1 & 2 & 3 & 4 \\    
    \hline
    AlexNet & 21.39 & 27.23 & 31.08 & 34.84 & 45.09 \\
    \hline
    Ours w/o context & 28.97 & 36.32 & \textbf{40.03} & \textbf{43.03} & 51.34 \\
    \hline
    Ours w/ context &  \textbf{30.13} & \textbf{36.38} & 39.96 & 42.90 & \textbf{51.74} \\
    \hline
\end{tabular}
\end{center}
\caption{Quantitative results for \textbf{pouring prediction}. The results for different edit distances of the groundtruth and predicted sequences are shown.}
\vspace{-0.5cm}
\label{tab:pourpred}
\end{table}

\section{Conclusion}
Reasoning about containers and the behavior of the liquids inside them is an important component of visual reasoning. However, it has not received much attention in the computer vision community. In this paper, we focused on four different tasks in this area, where the inference relies only on a single RGB image: (1) volume estimation, (2) content estimation, (3) comparative volume estimation, and (4) pouring prediction. We introduced the COQE dataset to train and evaluate our models. In the future, we plan to consider liquid attributes such as viscosity for more accurate prediction of pouring. Moreover, we plan to incorporate other modalities so we can perform more sophisticated reasoning in scenarios that the visual cues alone are not enough (e.g., opaque containers). %incorporate more sophisticated forms of contextual information such as depth or surface normals.

\noindent\textbf{Acknowledgements:} This work is in part supported by ONR N00014-13-1-0720, NSF IIS-1338054, NSF NRI-1637479, NSF IIS-1652052, NSF NRI-1525251, Allen Distinguished Investigator Award, and the Allen Institute for Artificial Intelligence. We would also like to thank Aaron Walsman for his help with preparing the 3D CAD models.

{\small
\bibliographystyle{ieee}
\bibliography{egbib}
}

\end{document}